\documentclass[10pt,twocolumn,letterpaper]{article}

\usepackage{cvpr}
\usepackage{times}
\usepackage{epsfig}
\usepackage{graphicx}
\usepackage{amsmath}
\usepackage{amssymb}
\usepackage{subcaption}
\usepackage{booktabs}
\usepackage{threeparttable}
\usepackage{tablefootnote}
\makeatletter
\@namedef{ver@everyshi.sty}{}
\makeatother
\usepackage{tikz}

\usepackage{tabularx,ragged2e}
\newcolumntype{C}{>{\raggedright\arraybackslash}X} 

\usepackage{float}

\newcommand{\todo}[1]{}
\renewcommand{\todo}[1]{{\color{red} TODO: {#1}}}

\usepackage[pagebackref=true,breaklinks=true,letterpaper=true,colorlinks,bookmarks=false]{hyperref}

\cvprfinalcopy 

\def\cvprPaperID{****} 

\begin{document}

\title{A Dataset and Benchmarks for Multimedia Social Analysis}

\author{Bofan Xue\\
University of California, Berkeley\\
{\tt\small xuebofan@berkeley.edu}
\and
David Chan\\
University of California, Berkeley\\
{\tt\small davidchan@berkeley.edu}
\and
John Canny\\
University of California, Berkeley\\
{\tt\small canny@berkeley.edu}
}

\maketitle
\thispagestyle{empty}
\pagestyle{empty}

\begin{abstract}
    We present a new publicly available dataset with the goal of advancing multi-modality learning by offering vision and language data within the same context. This is achieved by obtaining data from a social media website with posts containing multiple paired images/videos and text, along with comment trees containing images/videos and/or text. With a total of 677k posts, 2.9 million post images, 488k post videos, 1.4 million comment images, 4.6 million comment videos, and 96.9 million comments, data from different modalities can be jointly used to improve performances for a variety of tasks such as image captioning, image classification, next frame prediction, sentiment analysis, and language modeling. We present a wide range of statistics for our dataset. Finally, we provide baseline performance analysis for one of the regression tasks using pre-trained models and several fully connected networks. 
\end{abstract}

\section{Introduction}
    Generating an understanding of the world through multiple modalities remains an unsolved problem in artificial intelligence. While independent tasks such as automated video description \cite{ferraro2015survey}, image captioning \cite{sharma2018conceptual,vinyals2016show}, image classification and object detection \cite{liu2020deep}, scene graph detection \cite{del2013state} and others have all received significant individual attention, there has been relatively little work on large-scale multi-modal learning, which will be required for future intelligent systems. Indeed, all of the mentioned tasks use at most two modalities (With video description sometimes making use of audio features), while humans experience at least five modalities simultaneously. Building models which can understand and describe a context through multiple modalities have been explored in works such as Kaiser \etal \cite{kaiser2017one} and Tsai \etal \cite{tsai2018learning}, however these works suffer from a lack of paired data and must develop ways to train where paired inputs and outputs are not available for all samples. Learning on such out of sync data is notably different from the kinds of data that humans learn on, where we have synchronized audios, videos, and language information. 
    
    In this work, we introduce Social Vision and Language Dataset (SVLD), a new public dataset based on the \url{imgur.com} website, which has in-sync vision, language, and social data, along with some related metadata. This dataset consisting of over 677k posts, 2.9 million post images, 488k post videos, 1.4 million comment images, 4.6 million comment videos, and 96.9 million multi-modal comments all with paired reference metadata information, tags, and classification labels will help us to explore long-standing questions in multi-modal learning. While some datasets exist \cite{reddit,sharma2018conceptual,abu2016youtube,miech2019howto100m} containing online data (see section \ref{sec:related} for more info), none keep the same level of detail or information for each post at the same scale.

\section{Related Work}
\label{sec:related}
Vision and language research has been a long-standing problem in the AI community. Beginning with dual-modal paired datasets such as MSCOCO \cite{vinyals2016show}, which pairs captions and images, and extending to large-scale alt-text based captioning datasets such as Conceptual Captions \cite{sharma2018conceptual}, there has been a wealth of work on trying to explain images with text. Less work has been done on the same task in the video domain. Datasets such as MSR-VTT \cite{xu2016msr} and LSDMC \cite{rohrbach2017joint} have begun to explore professionally annotated video, while HowTo100M \cite{miech2019howto100m} is a large-scale video dataset that draws from internet-only annotations. Indeed, there seem to be two categories of datasets for description: small scale datasets that are professionally annotated, and large-scale internet datasets that are scraped and have only alt-text annotations. See \cite{ferraro2015survey} for a further overview of video and image description datasets.

While we are not the first group to collect a dataset of large-scale image samples (\cite{leskovec2016snap} and \cite{reddit}, both scrape from Reddit, a closely related site), current large scale internet datasets often only scrape a single modality. The SNAP dataset of Reddit posts \cite{leskovec2016snap} scrapes only the image, and the associated metadata (rating, title, and the number of comments). The large scale Reddit comment database \cite{reddit} is massive (containing 1.7 billion comments), but scrapes only comments and does not contain images/videos, nor images/videos in replies. Our dataset seeks to remedy this, by collecting in-sync data in multiple modalities, placing posts (which can be images/video), their associated comment trees (which can contain images/videos as comments) and metadata information such as the title, description, post score, and tag in a single accessible dataset.

In addition to large scale image and video captioning datasets, paired datasets such as RECOLA \cite{ringeval2013introducing}, IMEOCAP \cite{busso2008iemocap} and CMU-MOSI \cite{zadeh2018multimodal}, which have audio, video and text modalities have been proposed. These datasets have high-quality annotations, however are significantly smaller (each containing under 100K samples). Another modern multi-modal dataset is the Something-Something dataset \cite{goyal2017something}, which contains 200K short video clips paired with text data. The Something-Something dataset, while significant, is composed primarily of simple ideas and relationships and does not encode complex natural language challenges such as co-referencing and social relevance. Our proposed dataset has paired image, video, raw text and social data - which is easy to use for social analysis (such as predicting up-votes, see section \ref{sec:experiments}), or other understanding tasks such as captioning, or classification. 

A number of state of the art models for content understanding would benefit from the paired data that we have collected. Tsai \etal \cite{tsai2018learning} present a model for sentiment analysis which learns factorized representations for each modality, however must evaluate it on small-scale datasets \cite{park2014computational,wollmer2013youtube,zadeh2018multimodal,busso2008iemocap}, each with less than 1000 videos. Allowing such a model or similar models, such as those presented in \cite{liang2018multimodal} to be used on large-scale paired data, is an appealing avenue for the evaluation of the learned representations.

Another model proposed by Kaiser \etal uses MS-COCO \cite{vinyals2016show}, Imagenet \cite{deng2009imagenet}, the WSJ corpus \cite{paul1992design}, and the WMT translation corpus \cite{rej2017findings} to learn close to state of the art understanding models on the unpaired data. Such a model can be powerful without un-paired data, however it would be interesting to see how such a model can perform with limited paired data supervision, as could be provided by our dataset. 

\section{Dataset}

    SVLD is collected\footnote{The dataset was collected over the course of 8 months using the Imgur API and two open-source crawling tools \cite{imgur_scraper} and \cite{imgur_scraper_2}} from the social media website \textit{www.imgur.com}, a social media platform similar to Reddit, with a focus on socially relevant content and photo sharing. Our dataset is unique in that it innately combines both vision and language data within the same context--the social media format naturally gives rise to posts and responses that include both vision and language. SVLD contains approximately 680,000 total samples. A more detailed count of the number of samples, and some base size statistics are given in Table \ref{tab:raw_ds_statistics}.
    
    \subsection{What's in an example?}
    \label{sec:samples}
    
\begin{figure*}
    \centering
    \includegraphics[width=\textwidth]{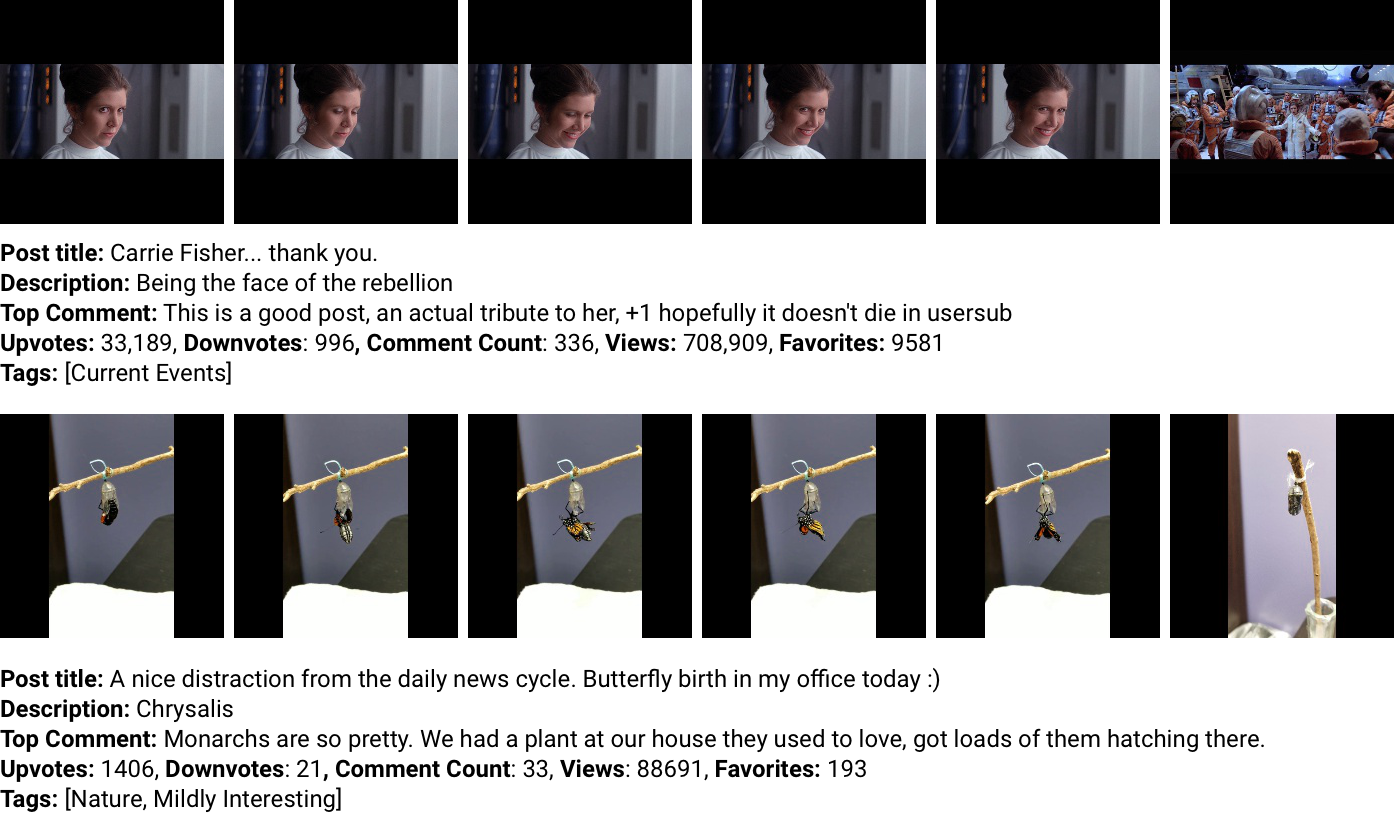}
    \caption{Some examples from the dataset. Here, we omit the comments and comment trees for clarity - for more information about comments, see section \ref{sec:comments} }
    \label{fig:examples}
    \end{figure*}

    Each sample in the dataset contains a mixture of images and video (some of which contain paired image/video descriptions), along with a natural language post title, a full set of comment trees, social information (such as the number of up-votes, down-votes, and favorites), and a set of post-classification tags. Some qualitative examples of the data that we have collected are given in Figure \ref{fig:examples}. In this section, we discuss in depth the actual content of each part and how popular multi-modal datasets differ in both scale and goal.
    
    \subsubsection{Images/Videos}
    
    Each sample in the dataset (corresponding to a post) contains a set of images and videos, ranging from a single image or GIF to large albums of images (related in context) or longer GIFs. This is a distinct difference from almost all vision datasets, which usually contain only a single modality \cite{vinyals2016show,sharma2018conceptual,deng2009imagenet,xu2016msr}, and usually do not contain local contextual grouping. Images and videos can be grouped in a post based on similar content, but also based on higher-level themes such as "funny", "mildly interesting", or "memes". These higher-level groupings have the potential to enable the additional study of visual data and transcend the issue of sticking with single temporal modalities.
    
    \subsubsection{Descriptions}
    
    In addition to having a collection of images and videos, each image and video in the dataset can be paired with an optional description. There is an extensive range of potential descriptions, ranging from comments on the individual image/video to summaries that place the image/video in the broader context of the post. 
    
    \subsubsection{Title}
    
    Each post contains a natural language title. Usually, this is a rich source of data detailing the contextual information about the entire post. Together with single image/video posts, titles can be used for image/video captioning and description tasks, as well as post-retrieval and search problems.
    
    \subsubsection{Tags}
    
    When creating a post, users are encouraged to tag their posts with a large set of tags. These tags range from basic classification to describing the higher-level social meaning. Unlike any clear-cut object classification dataset (such as ImageNet \cite{deng2009imagenet}), most of the tags in our dataset belong to the second type and describe higher-level social concepts. The most common tags in our dataset are: `funny', `aww', `memes', and `awesome'. In many cases, however, these higher-level tags are simultaneously joined with object-classification style tags, giving image classification models opportunities to learn the relationship between the two. In addition, these tags allow for learning social concepts that are one step beyond current classification, and perhaps even enabling conditional generative models tuned to social behavior.

    \subsubsection{Upvotes/Downvotes/Favorites}
    
    In addition to the tags, each post comes with upvotes, downvotes, and favorites as paired social information. These metrics are direct responses from users (who ``upvote" and ``favorite" things they like and ``downvote" things they dislike). This metric provides a highly unambiguous metric of which posts are of interest to the community, and which posts are expected to do well on the platform. Modeling this distribution is one of the key problems in social dynamics \cite{risch2020top}. 
    
    \subsubsection{Comments and Comment Trees}
    \label{sec:comments}
    
    In response to the post's content, each post has a forest of comment trees, where each comment can be text and/or image/GIF. These comment trees can have several different styles, such as opinion sharing (a shallow tree with lots of branches), Q\&A (a deep tree with limited branches), or general discussion (a mixture of both). There are interesting cases that are almost exclusively found in our dataset in that the user splits a giant text comment into smaller chunks and replies to the previous chunk with the next chunk, which improves readability and makes the entire comment more engaging. The wide range of comment tree styles, combined with the multi-modality nature of the comments, has the potential to enhance the modeling of contextual language, question answering, and social dynamics. 
    
    \subsubsection{Views \& Date}
    
    We also collect the number of views and the post date for each post (visualized in Figure \ref{fig:raw_dataset_distributions}). They provide us with an opportunity to examine the shifting social distributions over time, which is an interesting non-stationary modeling problem (See section \ref{sec:possible}).

    \subsection{Dataset Statistics}
    
\begin{figure*}
    \begin{subfigure}{\textwidth}
      \centering
      \includegraphics[width=1\textwidth]{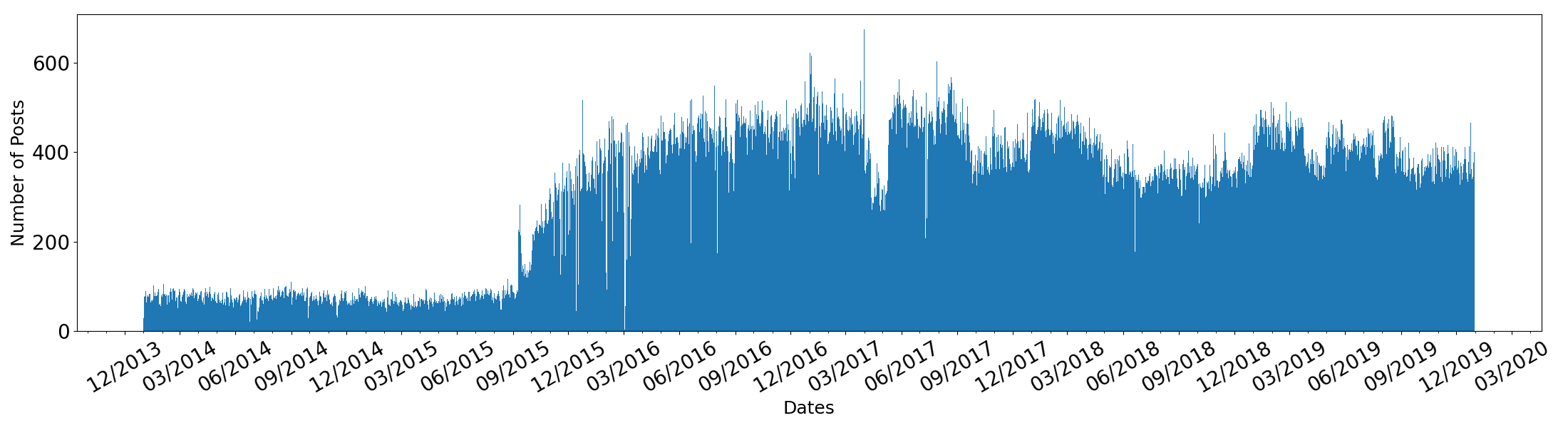}
    \end{subfigure}
    \begin{subfigure}{\textwidth}
      \centering
        \includegraphics[width=1\textwidth]{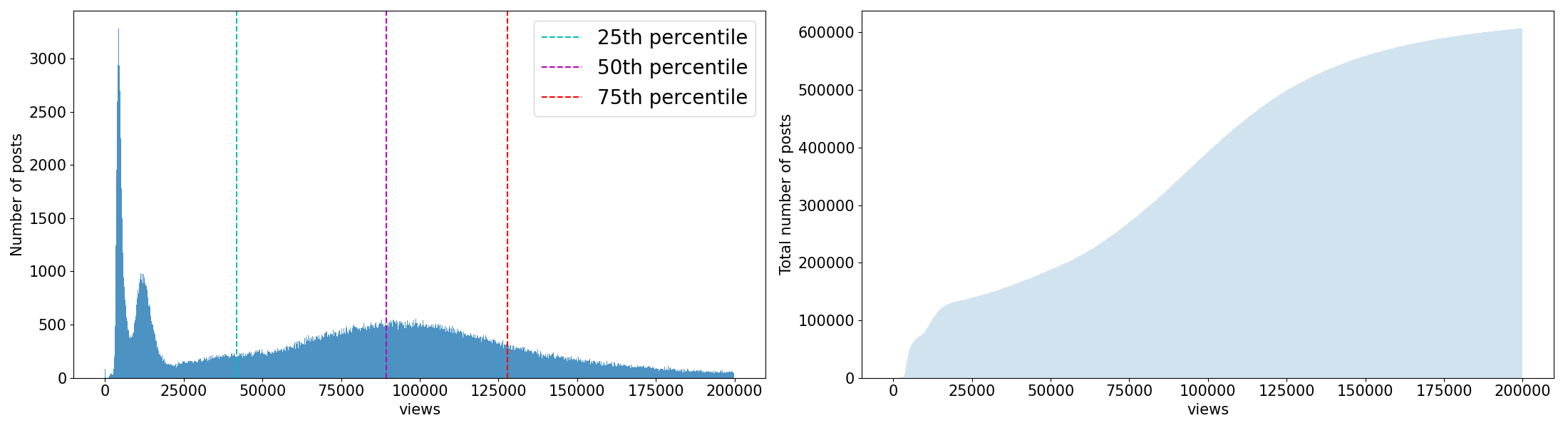}
    \end{subfigure}
    \caption{Top: Date distribution. We can see the shifting trends, and increasing popularity on the website, as it grows in relevance after 2015. Bottom: Views distribution for our dataset. Both bar plots contain 2000 bins with bin size 100. For rendering purposes, only posts with less than 200,000 views are shown, which includes 89.70\% of the posts.}
    \label{fig:raw_dataset_distributions}
    \end{figure*}

    Figure \ref{fig:raw_dataset_distributions} gives a distribution of dates of the collected samples. By collecting data at multiple points in time (as well as annotating the times that these posts were written), we have the ability to work with non-stationary social distributions and answer questions about how the paired vision and language distribution can evolve over time\footnote{The irregularities in terms of the number of posts per day are caused by various difficulties encountered during scraping, including Imgur server errors, network errors, incorrect website addresses, and errors from packages for processing images.}. 
    
    We also collect some statistics on the individual posts, given in Table \ref{tab:post_stats}. We can see from these single-post statistics that there is a substantial variation in the data contained in this ``in the wild" dataset, with posts containing collections of both image and video information. In addition, the bi-gram perplexity of the dataset is relatively high (in line with the WSJ NLP corpus \cite{paul1992design}), suggesting complex interactions in the dataset that cannot be learned from n-gram language models alone. 
    
    It's interesting to note that most of the collected data is overwhelmingly positive in terms of votes. With very high points ratio (upvotes / upvotes $+$ downvotes), we can see that most content is well received. This is an interesting imbalanced data problem, since detecting the few negative examples could significantly improve content moderation (as these examples have already been filtered by the Imgur content moderation tool).
    
    Figure \ref{fig:raw_dataset_distributions} also gives a distribution of the number of views (both cumulative and histogram-form). We can see that the number of views is largely long-tailed. Many posts have very few views, while there are additional peaks at approximately 10,000 and 1,000,000 views. The figures disregard a relatively long tail of the distribution, where 10\% of the data has more than 2M views.
    
    \begin{table}
        \centering
        \begin{tabular}{|l|l|}
        \hline
        \multicolumn{2}{|c|}{\textbf{Post statistics}} \\
        \hline
        Days included & 2191 \\
        Total number of posts & 677,181 \\
        \quad \textit{Posts with single image} & 308,222 \\
        \quad \textit{Posts with single video} & 215,295 \\
        \quad \textit{Posts with multiple images} & 113,619 \\
        \quad \textit{Posts with multiple videos} & 15,407 \\
        \quad \textit{Posts with both images and videos} & 24,638 \\
        Total number of post images & 2,907,478 \\
        Total number of post videos & 488,384 \\
        Total number of image/video descriptions & 917,833 \\
        \hline
        \multicolumn{2}{|c|}{\textbf{Comment statistics}} \\
        \hline
        Total number of comment images & 1,478,108 \\
        Total number of comment videos & 4,689,107 \\
        Total number of comment trees & 38,642,206 \\
        Total number of comment leaves & 63,826,049 \\
        Total number of comments & 96,961,858 \\
        \hline
        \end{tabular}
        \caption{Raw statistics for SVLD.}
        \label{tab:raw_ds_statistics}
    \end{table}

\begin{table}
\begin{minipage}{\linewidth}
\small
\begin{threeparttable}
\begin{tabular}{|c|c|c|}
    \hline
    \textbf{Statistic} & \textbf{Mean} & \textbf{Standard Deviation}\\
    \hline
    Number of images per post & 4.2935 & 16.3508\\
    Number of videos per post & 0.7212 & 3.5058\\
    Post comment tree depth & 1.5616 & 1.6974 \\
    Description length & 35.6132 & 101.9635 \\
    Description perplexity\tnote{1} & 49.9966 & 134.2269\\
    Title length & 8.1749 & 6.4226\\
    Title perplexity\tnote{1} & 111.5235 & 118.4065 \\
    Comment length & 16.6412 & 10.4310 \\
    Comment perplexity\tnote{1} & 111.9312 & 180.4790\\
    Point ratio & 0.9577 & 0.0364 \\
    \hline
    \textbf{Statistic} & \textbf{Value} &\\
    \hline
    Description vocab size & 28,349 & \\
    Title vocab size & 25,670 & \\
    Comment vocab size & 29,026 & \\
    \hline
\end{tabular}
\begin{tablenotes}
\item[1] Perplexity given is empirical bi-gram perplexity
\end{tablenotes}
\caption{Statistics over individual posts. The large unique word sizes and perplexities compared with datasets such as MSR-VTT, Something-Something, and MSCOCO suggest that this is a more faithful rendering of "in-the-wild" data, which real-world understanding systems will encounter.}
\label{tab:post_stats}
\end{threeparttable}
\end{minipage}
\end{table}

\section{Directions for Exploration}
\label{sec:possible}
   SVLD enables further exploration in many different fields and directions. In this section, we propose some potential directions for research enabled by this large-scale paired social dataset.  
   
\subsection{Multi-modal experiments}

    As seen in section \ref{sec:samples}, our dataset provides a large number of modalities. This allows for a matrix of cross-modality prediction experiments, ranging from well-studied tasks in social contexts such as predicting the number of upvotes/downvotes/views/tags from the rest of the post context (See Section \ref{sec:experiments}), to far more difficult modeling tasks such as selecting or predicting the top-comment in context of the post (which requires modeling complex social dynamics and commonsense reasoning) \cite{risch2020top} or user-modeling \cite{zhang2019deep}. In addition, there is little data fusing multiple visual modalities (images + video) and multiple natural language distributions (comments + titles + descriptions) in order to make inferences. This multiple-distribution context is much closer to what would be observed and modeled in real-world learning scenarios than single-modality, single-distribution data available in current datasets \cite{li2012literature}.
    
\subsection{Advanced models}
    
    Models such as "One model to Learn them all" \cite{kaiser2017one} have shown that large scale multi-modal training can lead to strong understandings of a joint linguistic space. Models for image and video captioning and analysis such as those in \cite{anderson2018bottom,yang2019auto,xu2019scene,ferraro2015survey} have taken the first steps towards large-scale multi-modal learning. We hope that this dataset can provide a platform for exploring more complex model architectures for multiple modes, and fusion beyond the simple baselines that we give in section \ref{sec:models}. Understanding how to fuse multiple modes of data for the same sample is a primary application for this dataset. This area of research has been relatively under-explored \cite{ramachandram2017deep}, and we hope that making access to a large-scale multi-modal dataset simple and accessible will help to spur research in this area. Given such a diverse set of possible tasks, ranging from modeling the tags to modeling the titles and top-comments, we can begin to leverage and benchmark fusion methods and explore what information is required to learn social metrics.
    
\subsection{Learning through comment trees}

    Datasets such as SNAP \cite{leskovec2016snap} and the Reddit comment dataset \cite{reddit} contain comment trees, however they often do not contain the associated context. SVLD is the first to provide a full comment tree with paired context information. In addition, SVLD also scrapes images/videos that appear as responses in the comments, which is not done by \cite{leskovec2016snap} and \cite{reddit}. From a social media and modeling perspective, this is extremely important. For example, trying to model a response chat-bot based solely on the comment tree without the associated context is an extremely challenging task. It is an open research question on how to effectively extract information from a tree of natural languages that extends beyond simple question-answering, not to mention how to learn from a forest of comment trees with different sizes that contain images, videos, and texts.
   
\subsection{Sets of images and videos with paired descriptions}
    Each post in the dataset may contain multiple images, videos, or a combination of them, and contain the description from none at all to one for each image/video. It is no doubt that, in a given post, any image or video and its description have their significance and can be used to predict tags, title, top comment, points ratio, or views, but from a broader context those predicted results should associate with the entire post, not with any single image, video, or description. Another possible avenue of future research would be to explore how to model the clustering algorithms that humans use to group images and videos into posts. Posts in the dataset are often semantically related in complex ways - perhaps they tell a story, or maybe they are linked by some higher theme. Training models to model these semantic relationships is an interesting avenue of research enabled by this data.
 
\subsection{Social theme shift through time}

    Another exciting and useful direction in modeling social dynamics is learning from non-stationary social theme shifts through time \cite{sayed2012learning}. With time stamp on each post, it is possible to observe on a macroscopic level variations in vocabulary, public opinions, or social topics through various time periods. These data may help to train a model that is capable of predicting future social themes. Furthermore, it may help to examine which kinds of posts are more prevalent on certain days, as with holidays or periods with special events. 
    
\subsection{Factors affecting social relevance}

    This is the most generic and the hardest question. \textit{Why does a given post have so many views, ups, or favorites? Why do some posts trigger vastly more comments than others? Why do some posts seem so engaging and meaningful for people?} Indeed, certain images and videos inherently bear the quality like funny, sad, or interesting and may help to partially explain the popularity for a given post. However, beyond a certain point, these factors themselves cannot explain high social engagement that post has. We hypothesize that post's relationship with contemporary news and specific events and/or several of the post's comments contributes to its high social relevance. With a multi-modality model that incorporates all available information, we believe it can fathom what makes a post socially engaging. This model, in turn, can help machines to behave more like humans, gain a better understanding of emotions, and create posts or comments that intrigues real humans. This model could help with the development of tools designed to produce relevant content, or perform content moderation at a large-scale level. 

\section{Baseline Social Analysis Experiments}
\label{sec:experiments}

One of the potential tasks that this novel dataset imagines is the prediction of the social score a particular post might receive. Such a tool has wide-reaching application from content moderation to advertising and social analytics \cite{stoddard2015popularity,rizos2016predicting,segall2012predicting,wang2017predicting}. Here, we present a basic first-look at the performance of such a tool on our dataset, and we show that there is significant room for research improvement over simple multi-modal baselines. 

Our goal in this baseline experiment is to model the points-ratio of a sample, the number of upvotes divided by the total number of votes. This is a number between zero and one, which roughly reflects the attitude of the community towards a particular post. 

\subsection{Data Cleaning \& Processing}
    
    In our selection to build a dataset subset for training, we want to ensure that each type of post, judged by its topic, theme, and social meaning, has enough candidates to allow model training to become possible. We selected 'tag' as the criterion, retaining posts that contain any of the top 1000 used tags out of 113,757 different tags. These 1000 tags cover 74.55\% of 1,145,595 tag\ instances throughout all posts. The dataset subset consists of 324,888 examples for training and 40,448 examples for validation. An unreleased partition is held off as the test set. 

    \textbf{Images:} We pre-process each image\footnote{Based on a single random image from each post} by resizing its longest dimension to 500 pixels, symmetrically padding zeros to 500 * 500, and downsizing to 224 * 224.

    \textbf{Videos:} To enable model training over videos, we sample at most 5 frames per second, then perform the same pre-processing described in the images section for each frame. If the video has less than 10 frames or, in edge-cases, if the video has 0 frames per second, all frames will be retained.  To enable batch training, we limit the number of frames for the video to 64 and pad with the last frame if the video does not have a sufficient number of frames. 
    
    \textbf{Description/Title/Top-Comment:} We use a standard BERT tokenizer \cite{Wolf2019HuggingFacesTS} to tokenize all descriptions, and computed unique words based on tokenized results, including the `[UNK]' token. We then add `[CLS]' to the start and `[SEP]' to the end of all descriptions.

\subsection{Models}
\label{sec:models}

    \begin{figure}
        \def\layersep{1.5cm}
        \begin{tikzpicture}[shorten >=1pt,->,draw=black!50, node distance=\layersep]
            \tikzstyle{every pin edge}=[<-,shorten <=1pt]
            \tikzstyle{neuron}=[circle,fill=black!25,minimum size=17pt,inner sep=0pt]
            \tikzstyle{encoder}=[neuron, fill=green!50];
            \tikzstyle{decoder}=[neuron, fill=red!50];
            \tikzstyle{multi-modal encoder}=[neuron, fill=blue!50];
            \tikzstyle{annot} = [text width=5em, text centered]
        
            \node[encoder, pin=left:Image] (E-1) at (0,-1) {};
            \node[encoder, pin=left:Video] (E-2) at (0,-2) {};
            \node[encoder, pin=left:Title] (E-3) at (0,-3) {};
            \node[encoder, pin=left:Description] (E-4) at (0,-4) {};
            \node[encoder, pin=left:Comment] (E-5) at (0,-5) {};
        
            \foreach \name / \y in {1,...,1}
                \path[yshift=0.0cm]
                    node[multi-modal encoder] (C-1) at (\layersep,-3 cm) {};
        
            \node[decoder,pin={[pin edge={->}]right:Points ratio}] (D-3) at (2 * \layersep,-3){};
        
            \foreach \source in {1,...,5}
                \foreach \dest in {1,...,1}
                    \path (E-\source) edge (C-\dest);
        
            \foreach \source in {1,...,1}
                \foreach \dest in {3,...,3}
                    \path (C-\source) edge (D-\dest);
        
            \node[annot,above of=C-1, node distance=1.5cm] (hl) {multi-modal encoder};
            \node[annot,above of=E-1, node distance=1cm] {Encoders};
            \node[annot,above of=D-3, node distance=1cm] {Decoders};
        \end{tikzpicture}
        \caption{Illustration of our model}
        \label{fig:model_plot}
    \end{figure}
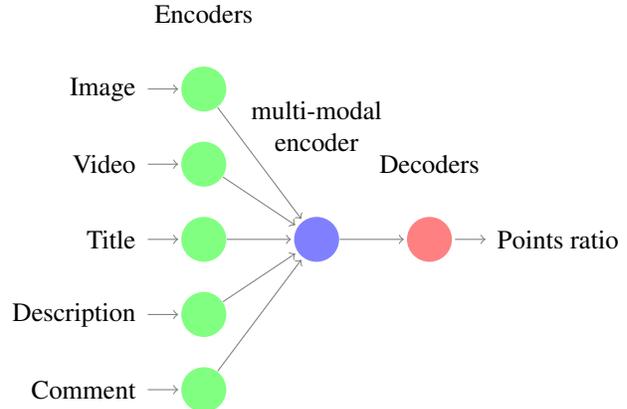

    To model the points ratio based on the multiple sets of input modalities, we suggest the model in Figure \ref{fig:model_plot}, inspired by the work by Kaiser \etal \cite{kaiser2017one} on multi-model encoders. The model extracts semantic information from each modality and uses a fusion layer to create a joint-embedding across all of the input modalities. 
    
    The model consists of three parts:
    
    \begin{itemize}
        \item Encoder modules that take in raw images, video frames, or texts and output corresponding embedding vector. Each modality has its own encoder.
        \item A multi-modal encoder that takes in embedding from all encoders and outputs a contextual embedding that contains information regarding each of input modalities.
        \item A decoder module that takes in contextual embedding and generates the labels - in our baseline experiment this corresponds to points-ratio, however in practice it could be any of the available modalities in the dataset.
    \end{itemize}
    
    \textbf{Image encoder:} We use a pre-trained ResNet152 over ImageNet dataset \cite{paszke2017automatic} as the feature extractor, collect its output before the last pooling layer, and feed the extracted features to a single layer fully connected network that outputs the final image embedding. 
    
    \textbf{Video encoder:} We use a pre-trained D3D network over Kinetics-600 dataset \cite{stroud2018d3d} as the feature extractor, collect its output before the last dropout layer, and feed the extracted features to a single layer fully connected network that outputs the final video embedding.

    \textbf{Language encoders:} For each of the language modalities, we instantiate an instance of the same encoder. The encoder first tokenizes the text via BERT large uncased tokenizer. It truncates the tokenized result to 100 tokens, adds `[CLS]' and `[SEP]' to the beginning and the end of the token sequence\footnote{This step is performed to ensure consistency with the pre-trained model}, and pads with `[UNK]' until the sequence has 102 tokens. The tokenized input then gets feed into the pre-trained BERT large uncased model \cite{Wolf2019HuggingFacesTS}, whose last layer embedding output serves as the input to a 3-layer attention-based encoder.

    \textbf{Multi-modal fusion:} We sequentially stack encoder embedding in the order of [image, video, description, title, top comment], skipping over any encoder that is not used. We use a two-layer fully connected network with hidden size 1024 to learn contextual embedding from encoder outputs.

    \textbf{Point ratio decoder:} The contextual embedding is first flattened across the encoder dimension, then gets passed into a single layer fully connected network to predict points ratios.

\subsection{Experiment Setup}

    \textbf{Loss:} Our goal is to predict the points-ratio of each of the samples, and we do so by optimizing the $L_2$ error between the predicted and target points ratio as a regression problem. 
    
    \textbf{Optimization:} The models are trained with the PyTorch Adam optimizer \cite{paszke2017automatic}, and learning rate $3e-4$ for 100 epochs with batch size of 128 on a single NVIDIA TITAN RTX GPU with an Intel i9-7960X CPU.
    
\section{Results \& Analysis}

    \begin{table}
    \begin{threeparttable}
    \small
    \begin{tabular}{@{}p{\linewidth}@{}}
    \centering
    \begin{tabular}{|c|c|}
    \hline
    \shortstack{\phantom{A} \\ \textbf{Method} \\ \phantom{A}} & \textbf{\shortstack{Mean $L_1$ \\ point ratio \\ prediction error}}\\
    \hline
    Baseline & $0.02330 \pm 0.02909$ \\
    Image Only\tnote{1} & $0.02293 \pm  0.03079$ \\
    Video Only &  $0.02111 \pm 0.02940$ \\
    Image + Video\tnote{1} & $0.02369 \pm 0.03073$ \\
    Description Only\tnote{2} & $0.02397 \pm 0.03180$ \\
    Title Only\tnote{4} & $0.02314 \pm 0.02882$ \\
    Top-Comment Only & $0.02578 \pm 0.03287$ \\
    \hline
    \shortstack{Description + Title + \\  Top-Comment\tnote{2}} & $0.02428 \pm 0.02824$ \\
    \hline
    All Encoders\tnote{3}\tnote{4} & $0.02323 \pm 0.03172$ \\
    \hline
    \end{tabular}
    \end{tabular}
    \begin{tablenotes}
    \small
    \item[1] $p$-value of Image vs. Image + Video is $0.547$
    \item[2] $p$-value of Description vs. Description + Title + Top-Comment is $0.597$
    \item[3] $p$-value of Video vs. All is $0.5754$
    \item[4] $p$-value of Title vs. All is $0.5701$
    \end{tablenotes}
    \caption{Validation performance of the baseline models. All values are pairwise statistically significant at $p=0.05$, with the exception of those mentioned in the footnotes.}
    \label{tab:results}
    \end{threeparttable}
    \end{table}

    Table \ref{tab:results} outlines the results of our baseline experimentation. Of the different methods for encoding, the image and video only encoders are the only ones that are able to outperform a simple mean baseline. While the results are a good first step towards modeling over the new data, they represent significant room for improvement beyond the current model behavior. 
    
    We expect that the main reason for the performance discrepancy in performance is the domain shift from the pre-trained models that we employed to the current data. The BERT models are trained on large-scale text data, and the image/video models are trained on ImageNet/Kinetics, which likely have a very different distribution of inputs than what is seen in the current data. By fine-tuning the image representations, and learning semi-supervised and unsupervised embeddings on the scale of our data, we expect that we can get far better performance. 
    
    In addition, it is possible that the large variation in performance is caused primarily by the ability of models to distinguish fine-grained differences in the points ratio between the examples. Because we originally trained the models with an L2 loss, the diminishing returns as we approach zero interact poorly with the optimization, leaving us a numerically unstable optimization problem. A next step would be to investigate Huber style losses, which remain differentiable close to zero but have behavior closer to L1 metric losses.

\section{Concluding Remarks}
    In this work we presented SVLD, a multi-modality dataset that innately combines both vision and language data in the context of social media, which has several key characteristics: it has 2.9 million images in posts and 1.4 million images in comments, 488k videos in posts and 4.6 million videos in comments, and a total of 96.9 million comments. We provide a wide range of statistics for the dataset as a whole and each of the modalities. Finally, we discuss several possible venues for future research and perform a simple baseline regression experiment. The dataset and baseline code for the processing of data and our experiments are publicly available at \url{https://cannylab.github.io/svld}.

{\small
\bibliographystyle{ieee_fullname}
\bibliography{egbib}

\begin{thebibliography}{10}
\bibitem{reddit}
r/datasets - i have every publicly available reddit comment for research. ~ 1.7
  billion comments @ 250 gb compressed. any interest in this?
\end{thebibliography}
}

\end{document}